\documentclass[10pt,twocolumn,letterpaper]{article}

\usepackage{cvpr}
\usepackage{times}
\usepackage{epsfig}
\usepackage{graphicx}
\usepackage{amsmath}
\usepackage{amssymb}
\usepackage{booktabs}
\usepackage{multirow}
\usepackage{graphicx}
\usepackage{subfigure}
\usepackage{enumitem}

\usepackage[pagebackref=true,breaklinks=true,letterpaper=true,colorlinks,bookmarks=false]{hyperref}

\cvprfinalcopy 

\ifcvprfinal\fi
\begin{document}

\title{OpenGF: An Ultra-Large-Scale Ground Filtering Dataset Built Upon Open ALS Point Clouds Around the World}


\author{Nannan Qin\textsuperscript{1,2}, Weikai Tan\textsuperscript{2}, Lingfei Ma\textsuperscript{3}, Dedong Zhang\textsuperscript{4}, Jonathan Li\textsuperscript{2,4*}\\
\textsuperscript{1}Key Laboratory of Planetary Sciences, Purple Mountain Observatory, \\
Chinese Academy of Sciences, Nanjing, JS 210023, China\\
\textsuperscript{2}Department of Geography and Environmental Management, \\
University of Waterloo, Waterloo, ON N2L 3G1 Canada\\
\textsuperscript{3}School of Statistics and Mathematics, Central University of Finance and Economics, Beijing 102206, China\\
\textsuperscript{4}Department of Systems Design Engineering, 
University of Waterloo, Waterloo, ON N2L 3G1 Canada\\
{\tt\small \{nannan.qin, weikai.tan, l53ma, dedong.zhang, junli\}@uwaterloo.ca}}

\maketitle

\begin{abstract}
Ground filtering has remained a widely studied but incompletely resolved bottleneck for decades in the automatic generation of high-precision digital elevation model, due to the dramatic changes of topography and the complex structures of objects. The recent breakthrough of supervised deep learning algorithms in 3D scene understanding brings new solutions for better solving such problems. However, there are few large-scale and scene-rich public datasets dedicated to ground extraction, which considerably limits the development of effective deep-learning-based ground filtering methods. To this end, we present OpenGF, first Ultra-Large-Scale Ground Filtering dataset covering over 47 $km^2$ of 9 different typical terrain scenes built upon open ALS point clouds of 4 different countries around the world. OpenGF contains more than half a billion finely labeled ground and non-ground points, thousands of times the number of labeled points than the de facto standard ISPRS filtertest dataset. We extensively evaluate the performance of state-of-the-art rule-based algorithms and 3D semantic segmentation networks on our dataset and provide a comprehensive analysis. The results have confirmed the capability of OpenGF to train  deep learning models effectively. This dataset is released 
at \url{https://github.com/Nathan-UW/OpenGF} to promote more advancing research for ground filtering and large-scale 3D geographic environment understanding.
\end{abstract}

\section{Introduction}

High-precision Digital Elevation Model (DEM) is an indispensable fundamental data for a variety of applications, including road survey and design, urban flood risk estimation~\cite{wang2018integrated}, carbon storage estimation~\cite{patenaude2004quantifying}, archaeology~\cite{canuto2018ancient}, deep space exploration~\cite{sugita2019geomorphology}, \etc. Airborne Laser Scanning (ALS) has been widely used to produce high-precision DEM, due to its capability of penetration through vegetation and efficient acquisition of highly precise and dense point clouds in large-scale environments with changing terrains. The essential cornerstone of the DEM generation from ALS data is ground filtering (GF), which is the process of accurately separating ground (GR) points from non-ground (NG) points.

\begin{figure}[t]
    \centering
    \includegraphics[width=\linewidth]{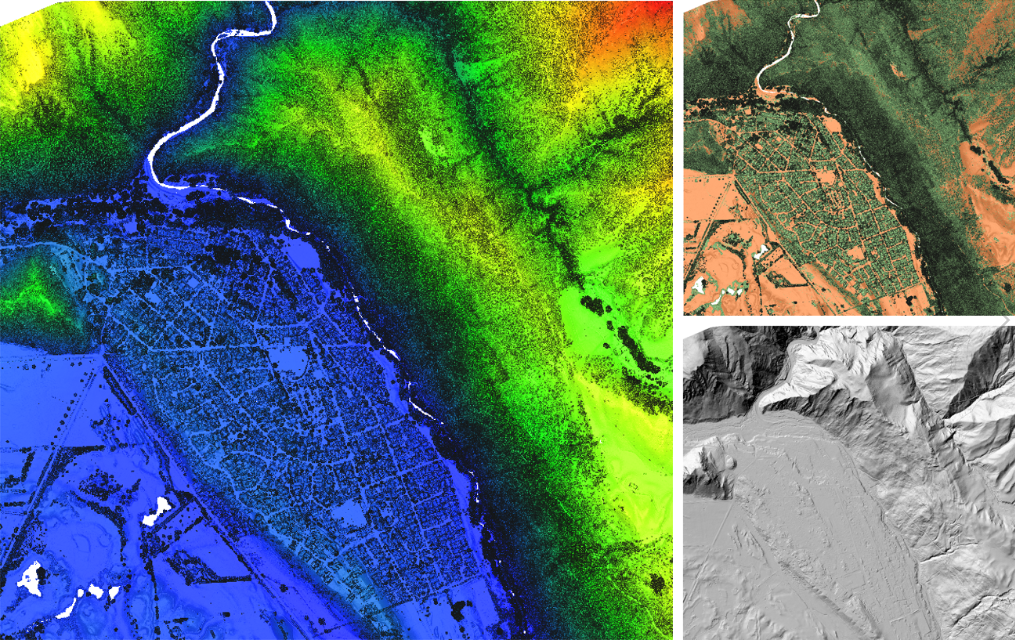}
    \caption{An example of the OpenGF dataset. Left: The original point cloud, Top right: Finely labeled GR (orange color) and NG (green color) points, Bottom right: DEM generated from the labeled GR points.}
    \label{fig:example}
\end{figure}

Due to significant topographic relief variations as well as the complex structures of natural or man-made land covers, notable intra-class differences in GR or NG class (\eg, plain vs steep slope, power line vs roof) are often observed on earth surfaces. Meanwhile, there are very similar geometric structures in these two categories (\eg, square flat ground vs large roof, cliff vs building facade). High intra-class differences and inter-class similarities make it extremely challenging to finely separate GR and NG points in the noise-contaminated point clouds. Therefore, despite having been studied for nearly three decades, the problem of GF has not yet been completely solved.

With the rapid development of deep learning techniques, several studies have applied them to GF, and made a breakthrough in the level of automation and accuracy compared with traditional rule-based methods. However, the large-scale datasets used in these deep-learning-based studies are rarely public access or lack diversity, and the widely used ISPRS test set\textsuperscript{\ref{filtertest}} for GF contains too few labeled samples. Therefore, there is an urgent demand for a new large-scale public dataset to promote the development of advanced deep learning algorithms for GF. 

The research object of GF is the changeable topography, which requires the dataset should not only contain a large number of training samples, but also cover as diverse terrain scenes as possible. However, it is far from straightforward  to finely separate GR points from NG points manually, especially in mountainous areas with randomly distorted terrain and dense vegetation. Thus, it is time-consuming and labor-consuming to build a high-quality GF dataset.

To successfully deal with the problem above, this paper builds an ultra-large-scale dataset for GF efficiently by taking advantage of the existing open ALS point clouds with clean bare earth class around the world. This dataset, called OpenGF, covers more than 47 $km^2$ and contains 9 different terrain scenes. It consists of a training set with diverse samples and a test set (one sample of the test set is shown in Figure~\ref{fig:example}) carefully selected from different regions. In addition, we conduct a comparative analysis of the performance of state-of-the-art rule-based algorithms and 3D semantic segmentation networks on OpenGF. The main contributions of this paper are as follows:

\begin{itemize}[leftmargin=*]
\setlength{\itemsep}{0pt}
\setlength{\parsep}{0pt}
\setlength{\parskip}{0pt}
  \item  Presenting an ultra-large-scale ALS point cloud dataset with both finely labeled GR points and diverse terrain scenes, dedicated to developing advanced deep learning methods for GF. To the best of our knowledge, no such datasets have been previously released. 
  \item Proposing a detailed strategy to quickly construct a large-scale dataset by leveraging existing open data around the world, which provides a new idea for the rapid construction of other large-scale datasets.
  \item Comparing and analyzing state-of-the-art rule-based and deep-learning-based algorithms on OpenGF, which verifies the effectiveness of our dataset and provides a common benchmark.
  \item Converting the traditional evaluation metrics of GF to the  equivalent  or  similar  metrics  widely  used  in  current classification tasks. To the best of our knowledge, such a proposal has not been made before.
\end{itemize}

\section{Related Point Cloud Datasets}
With the rapid development of 3D geographic data acquisition techniques, more and more outdoor large-scale point cloud datasets have been publicly released to support the deep-learning-based applications of geospatial data intelligence. 

According to different data acquisition methods, existing geospatial datasets for semantic labeling can be broadly classified into three categories: (1) TLS/MLS point cloud datasets. These datasets are usually used for outdoor roadway scene understanding, such as Semantic3D~\cite{hackel2017isprs}, Paris-Lille-3D~\cite{2017Paris}, SemanticKITTI~\cite{behley2019semantickitti}, Toronto-3D~\cite{tan2020toronto}. Although their sample sizes are very large, the geographic coverage of these datasets is still too small to be used for large-scale DEM extraction.  (2) Photogrammetric point cloud datasets, including Campus 3D~\cite{li2020campus3d}, 
SensatUrban~\cite{hu2020towards}. These datasets lack ground points in vegetated areas, because photogrammetry can't obtain ground points under vegetation. As a result, deep learning models trained on photogrammetric point cloud datasets have limitations in DEM extraction in vegetation areas. (3) ALS point cloud datasets. The majority of these datasets are specifically built for 3D urban classification, such as the ISPRS 3D semantic labeling dataset~\cite{2014Contextual}, DublinCity~\cite{DBLP:conf/bmvc/ZolanvariRRCSRS19}, DALES~\cite{varney2020dales}, LASDU~\cite{ye2020lasdu}. These datasets mainly focus on distinguishing as many objects of interest as possible (\eg, impervious surfaces, power lines, cars), so that they either didn't annotate bare earth points carefully (\eg, directly label low object points as ground for convenience), or lack samples of common non-urban terrain scenes (\eg, wilderness, mountains), which makes them can't well meet the needs of GF research.

\begin{table*}[t]
\centering
\begin{tabular*}{\textwidth}{c@{\extracolsep{\fill}}cccccc}
\toprule \midrule
\textbf{Dataset} &
  \textbf{Year} &
  \textbf{Coverage} &
  \textbf{Points} &
  \textbf{RGB} &
  \textbf{Collection} &
  \textbf{Application} \\ \midrule
Semantic3D \cite{hackel2017isprs} &
  2017 &
  - &
  4,000M &
  Yes &
  TLS &
  \multirow{4}{*}{\begin{tabular}[c]{@{}c@{}}Roadway-level \\ Semantic segmentation\end{tabular}} \\
Paris-Lille-3D~\cite{2017Paris}         & 2018 & $1940m$            & 143M  & No  & MLS                &  \\
SemanticKITTI~\cite{behley2019semantickitti}          & 2019 & $39200m$           & 4,549M & No  & MLS                &  \\
Toronto-3D~\cite{tan2020toronto}             & 2020 & $1000m$            & 78M   & Yes & MLS                &  \\ \midrule
Campus3D~\cite{li2020campus3d} &
  2020 &
  $1.6\times10^6m^2$ &
  937.1M &
  Yes &
  Photogrammetry &
  \multirow{6}{*}{\begin{tabular}[c]{@{}c@{}}Urban-level \\ Semantic segmentation\end{tabular}} \\
SensatUrban~\cite{hu2020towards}        & 2020 & $7.6\times10^6m^2$ & 2,847M & Yes & Photogrammetry &  \\
Vaihingen (ISPRS)~\cite{2014Contextual}                 & 2012 & -                  & 1.2M  & No  & ALS                &  \\
DublinCity~\cite{DBLP:conf/bmvc/ZolanvariRRCSRS19}            & 2019 & $2\times10^6m^2$   & 260M  & No  & ALS                &  \\
DALES~\cite{varney2020dales}                 & 2020 & $10\times10^6m^2$  & 505M  & No  & ALS                &  \\
LASDU~\cite{ye2020lasdu}                  & 2020 & $1\times10^6m^2$   & 3.1M  & No  & ALS                &  \\ \midrule
Filtertest (reference data)\textsuperscript{\ref{filtertest}} &
  - & $1.1\times10^6m^2$ 
   & 0.4M
   &
  No &
  ALS &
  \multirow{2}{*}{\begin{tabular}[c]{@{}c@{}}Land-level \\ Ground filtering\end{tabular}} \\
\textbf{OpenGF (ours)} & 2021 & $47.7\times10^6m^2$  & 542.1M  & No  & ALS   & \\ \bottomrule
\end{tabular*}  
\vspace{0.05cm}
\caption{Comparison with the representative 3D geospatial datasets for semantic labeling}
\label{tbl:datasets}
\end{table*}

Currently, there are two kinds of datasets used for testing GF algorithms. One is the well-known Filtertest dataset\footnote{\url{https://www.itc.nl/isprs/wgIII-3/filtertest}\label{filtertest}}, which was provided by the ISPRS before 2003 to compare the performance of various rule-based GF methods. In this dataset, 15 samples with different characteristics have been compiled by semi-automatic filtering and manual editing. However, there are only tens of thousands of points per sample, which is too few to be used to train supervised deep neural networks. The other kind is the self-used large-scale experimental data with ground truth in several recent articles on deep-learning-based methods, such as the South China Point Clouds used by Hu \etal~\cite{Hu_2016} , and the Point Clouds in Forested Environments used by Jin \etal~\cite{Jin_2020}. However, these datasets are either not publicly available or limited to a single scene and GR annotations of low quality.  

The development of effective deep-learning-based GF algorithms demands large-scale datasets with both fine-labeled GR and diverse terrain scenes. To improve the intelligent level of ground filtering of point clouds spanning over ultra large areas, OpenGF is first introduced in this paper to encourage developing creative deep learning models. Table~\ref{tbl:datasets} shows a comparison of comprehensive indicators of the above-mentioned representative datasets.

\section{First Ultra-Large-Scale Dataset for Ground Filtering: OpenGF}
Unlike most 3D understanding tasks, the objective of GF is to accurately extract GR points in various landforms, which means a high-quality GF dataset should have not only a wide coverage but also contain diverse terrain scenes. However, obtaining large-scale ALS point clouds, especially from different countries, is extremely expensive and difficult. Meanwhile, it is time-consuming and labor-consuming to produce high-quality manual labels of GR and NG in diverse terrain scenes. Therefore, it is challenging to build a large-scale dataset for GF only by a single research institution or team.

\begin{figure*}[t]
    \centering
    \includegraphics[width=\linewidth]{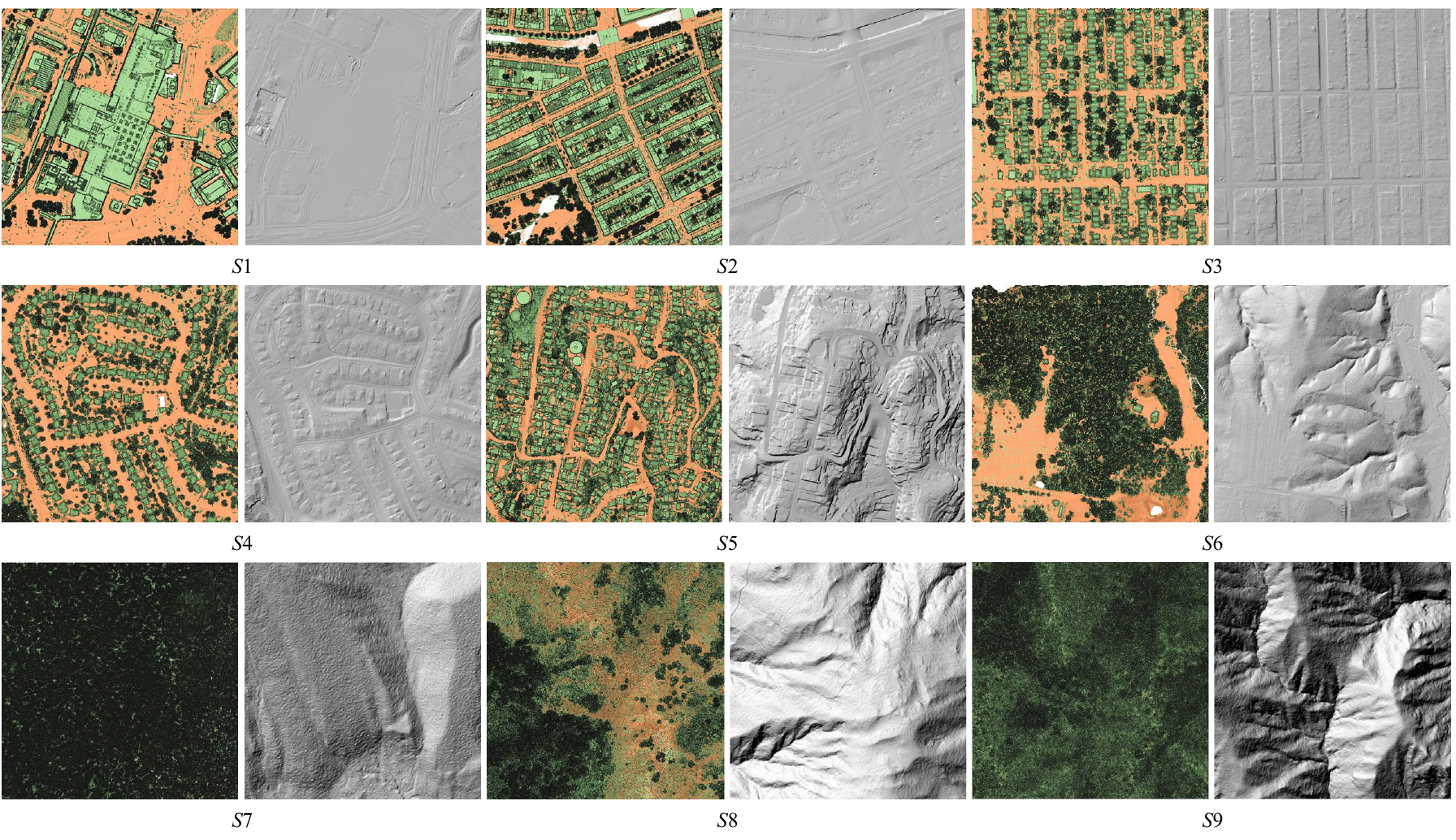}
    \caption{Examples of the validation split: nine point cloud tiles belong to the defined nine different terrain scenes. Each sample is displayed in pairs: the labeled point cloud on the left and the corresponding DEM on the right.}
    \label{fig:validation}
\end{figure*}

Fortunately, more and more ALS point clouds around the world have become available, and some of them already have fine GR annotation. In this case, unlike most 3D datasets, our ultra-large-scale dataset is quickly built by the leverage of existing open ALS data. The specific steps are as follows.

\begin{figure}[t]
\centering
\subfigure[The geographical distribution of the training set. Note that the world base map comes from OpenTopography\textsuperscript{\ref{opentopography}}.]{
\begin{minipage}[b]{0.48\textwidth}
\centering
\includegraphics[width=\textwidth]{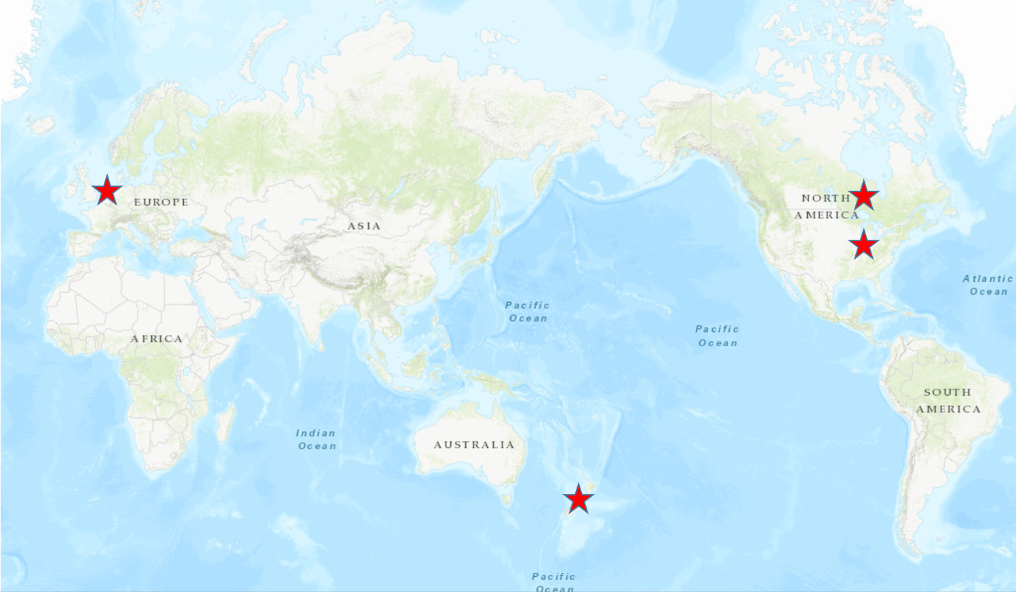} \label{fig:statistics_a}
\end{minipage}
}
\subfigure[The distribution of GR and NG points in different terrain scenes.]{
\begin{minipage}[b]{0.48\textwidth}
\centering
\includegraphics[width=\textwidth]{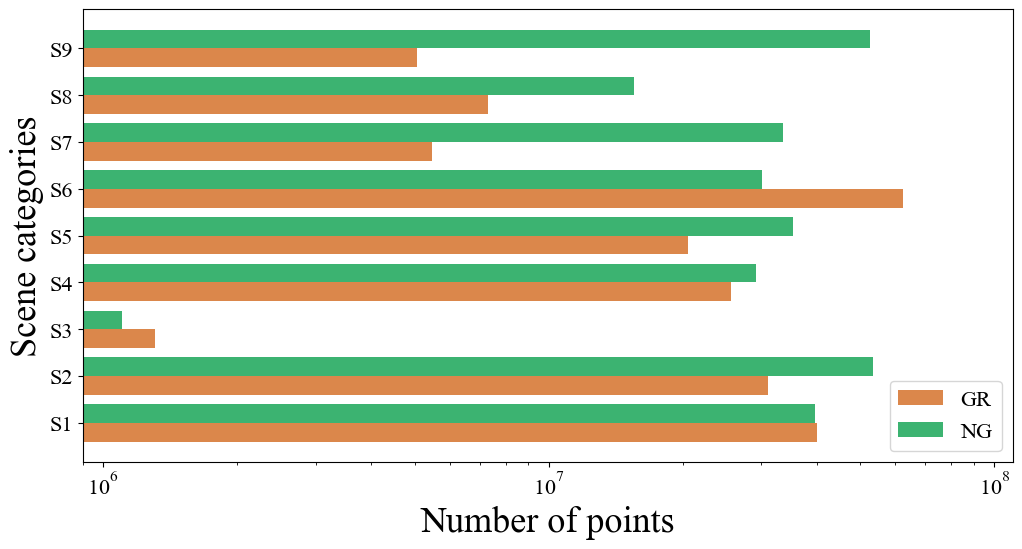} \label{fig:statistics_b}
\end{minipage}
}
\caption{Statistics of the training set. }
\label{fig:statistics}
\end{figure}

\subsection{Initial Point Clouds Selection}

Although the GR points in most open ALS point cloud data have been labeled, the quality of these annotations is inconsistent. To this end, we first pick out those point clouds with a high-quality GR category. Then, to avoid over-fitting, the picked point clouds are further screened and cross-referenced with satellite imagery to cover 4 prime terrain scenes as follow:

\begin{itemize}[leftmargin=*]
\setlength{\itemsep}{0pt}
\setlength{\parsep}{0pt}
\setlength{\parskip}{0pt}
  \item \textit{Metropolis}: Flat ground with dense or very large buildings.
  \item \textit{Small City}: Flat or rugged ground with many middle-size buildings.
  \item \textit{Village}: Natural ground with a few scattered buildings.
  \item \textit{Mountain}: Slope ground with dense or sparse vegetation.
\end{itemize}

Finally, two or three large-scale point clouds with fine GR labels in each typical terrain scene are extracted from the open ALS data of AHN3\footnote{\url{https://downloads.pdok.nl/ahn3-downloadpage/}}, OpenTopography\footnote{\url{https://portal.opentopography.org/datasets}\label{opentopography}}~\cite{LINZ2019}\cite{ACLINZ2019}\cite{WCCLINZ2020}\cite{TDCLINZ2020}\cite{IndianaMap2012}\cite{MarkLorang2013}, and Ontario Point Cloud (Lidar-Derived)\footnote{\url{https://geohub.lio.gov.on.ca/datasets/adf19376eecd4440a4579a73abe490f5}}. The descriptions of the extracted point clouds can be found from the corresponding official websites.

\subsection{Point Cloud Partition and Fine Picking}

To apply powerful GPU for training and testing, the extracted large-scale point clouds are segmented into tiles with the size of 500 $\times$ 500 $m^2$. Meanwhile, to keep each tile distinctly unique, there is no overlap between adjacent tiles.

After partition, 40 tiles with fine GR labels are selected from each terrain scene to ensure a sufficient and balanced number of samples. In particular, to contain as many terrain scenes as possible, inspired by previous research on terrain scene recognition~\cite{Qin_2018}, we try our best to pick out tiles belonging to different sub-scenes. Specifically, 20 tiles with large roofs ($S1$) and 20 tiles with dense roofs ($S2$) are picked out from metropolis areas. 10 tiles with flat ground ($S3$), 20 tiles with local undulating ground ($S4$), and 10 tiles with rugged ground ($S5$) are selected from small city areas. 40 tiles with scattered buildings ($S6$) are selected from village areas. 10 tiles with the gentle slope and dense vegetation ($S7$), 10 tiles with steep slope and sparse vegetation ($S8$), and 20 tiles with the steep slope and dense vegetation ($S9$) are selected from mountain areas. 

Finally, 4 prime terrain scenes are expanded further into 9 more detailed terrain scenes.  

\begin{figure}
\centering
\subfigure[Representative local areas of Test \uppercase\expandafter{\romannumeral1}. The red and blue lines indicate two different cross-section positions, and the corresponding points within the cross-sections are shown in the boxes with the same color.]{
\begin{minipage}[t]{0.48\textwidth}
\includegraphics[width=\textwidth]{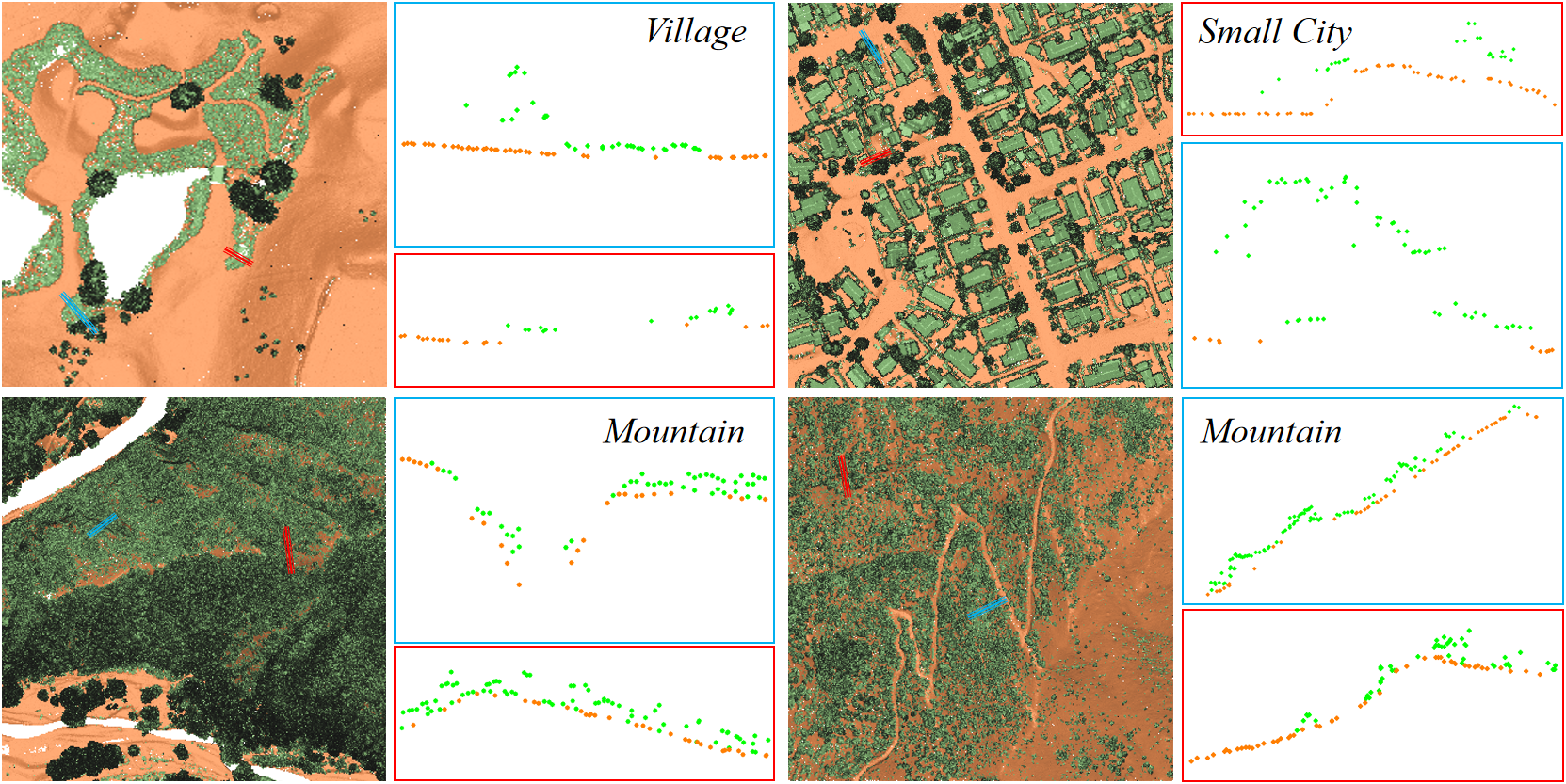} \label{fig:challenges_a} \\
\end{minipage}
}
\subfigure[The side view and four representative local areas of Test \uppercase\expandafter{\romannumeral2} are shown in the top and bottom, respectively. Red points represent outliers.]{
\begin{minipage}[t]{0.48\textwidth}
\includegraphics[width=\textwidth]{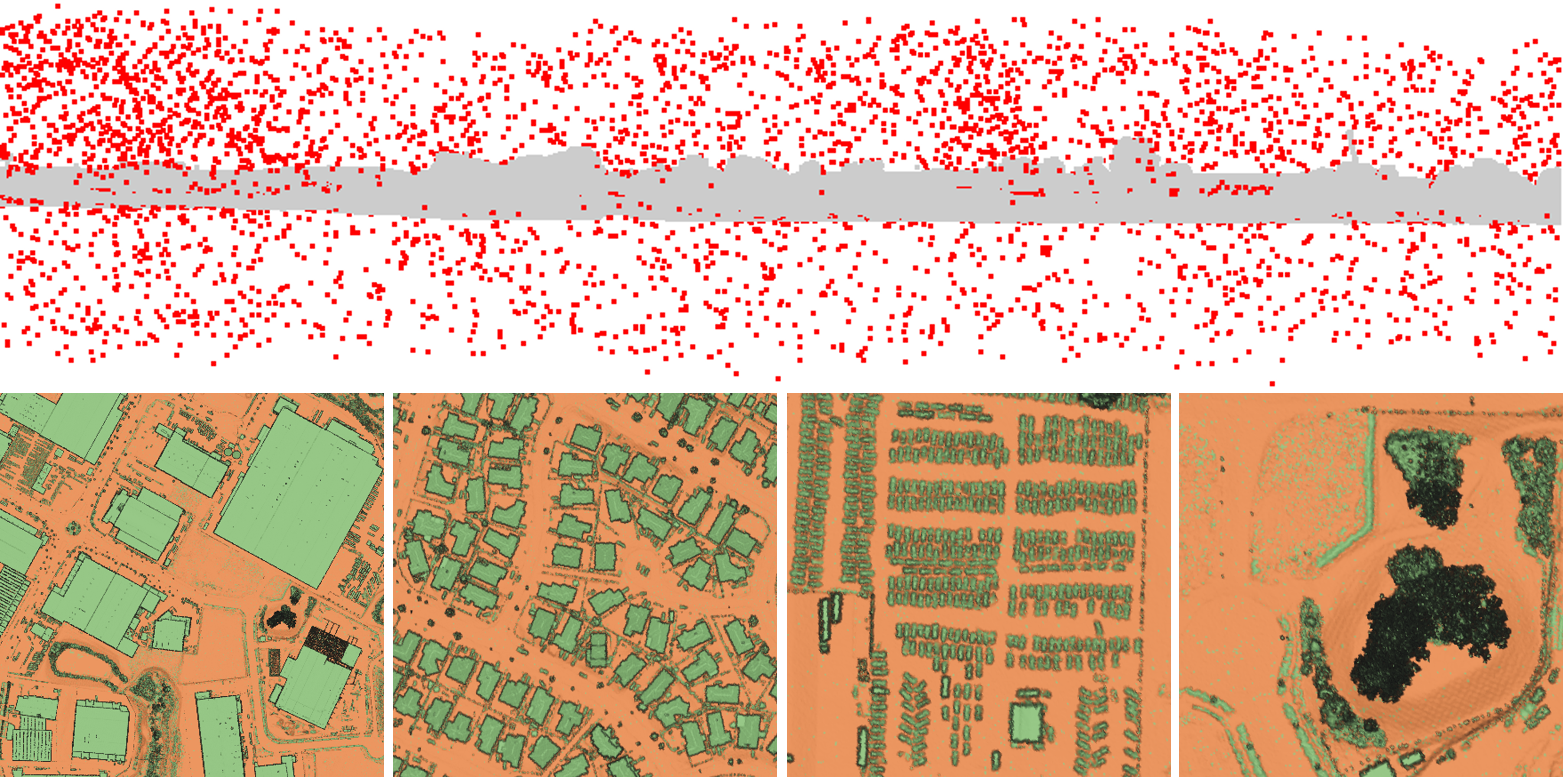} \label{fig:challenges_b}\\
\end{minipage}
}
\caption{Challenges of the test set.}
\label{fig:challenges}
\end{figure}

\subsection{Labeling of GR and NG Points}

Since we focus on GF, only two categories (i.e.,  GR and NG) are considered in this paper despite most of the point cloud tiles we selected contain more detailed categories, such as buildings, vegetation, bridges, water, overlapping, low and high outliers, \etc. To this end, the overlapping points and water surface points containing vegetation are firstly removed because they contain both NG and GR points. Then, the remaining classes are combined into NG, GR, and Unclassified. The definitions of these three classes are as follows:

\begin{itemize}[leftmargin=*]
\setlength{\itemsep}{0pt}
\setlength{\parsep}{0pt}
\setlength{\parskip}{0pt}
  \item \textit{Unclassified (class 0)}: Low outliers, high outliers.
  \item \textit{NG (class 1)}: Buildings, low vegetation, medium vegetation, high vegetation, bridges, cars, and other unclassified objects.
  \item \textit{GR (class 2)}: Bare earth, clean water surfaces.
\end{itemize}

 Note that outliers are assigned to one single category (\ie, Unclassified) in our dataset, so that researchers can decide whether to merge outliers into NG according to their specific needs.  Finally, the three combined categories are manually cross-checked to ensure consistency and high quality. It only totally takes several hours to complete the annotation of all points in our dataset.

\subsection{Statistics of Training Set}

The 160 point cloud tiles processed above are used as our training set, from which 9 representative samples (shown in Figure~\ref{fig:validation}) are carefully selected for validation. 

Figure~\ref{fig:statistics_a} shows that point cloud tiles are distributed in 4 different countries around the world, including Netherlands in Europe, New Zealand in Oceania, the United States and Canada in North America. The geographic distribution of the training set is shown with red stars. The scattered distribution enhances the diversity of our training samples.

Figure~\ref{fig:statistics_b} shows the distribution of GR and NG points in each terrain scene defined in this paper. From Figure~\ref{fig:statistics_b}, we can see that the distribution of class GR and NG is unbalanced in certain scenes. In particular, there are far more NG than GR points in mountain areas (including S7, S8, and S9), while village areas (S6) contain significantly more GR than NG points. In addition, the total number of points in S3 is far less than that in S5, although the coverage of the two scenes is the same, which indicates that the point density of our training set is diversified.

\subsection{Choosing of Challenging Test Set } 

To fully evaluate the generalization ability of deep neural networks, two challenging point clouds are extracted in areas different from the training set and processed as our test set.

Test \uppercase\expandafter{\romannumeral1}, one sample of our test set, is a challenging point cloud covering about 6.6 $km^2$ in a mixed area crossing village, small city, and mountain. The average point density of Test \uppercase\expandafter{\romannumeral1} is around 6 points/$m^2$. Four local areas of Test \uppercase\expandafter{\romannumeral1}, belonging to different scenes, are displayed in Figure~\ref{fig:challenges_a}. Based on the cross-section points shown in Figure~\ref{fig:challenges_a}, we can see that it is very difficult to distinguish the GR and NG points in certain local areas of Test \uppercase\expandafter{\romannumeral1}.

Test \uppercase\expandafter{\romannumeral2} is another point cloud spanning about 1.1 $km^2$ in metropolis areas. The average point density of Test \uppercase\expandafter{\romannumeral2} is about 14 points/$m^2$. As shown in Figure~\ref{fig:challenges_b}, Test \uppercase\expandafter{\romannumeral2} contains many low and high outliers and various sizes of objects (e.g., Large roofs, small roofs, cars, and grass). In particular, there is a large building whose length and width are both over 200 meters, which brings challenges to many state-of-the-art 3D semantic segmentation networks originally designed for small-scale point clouds. 

\section{Benchmark}

\subsection{Existing Ground Filtering Methods}

In general, existing ground filtering approaches can be divided into two categories: (1) \textit{Rule-based methods}. These conventional methods are generally based on artificially designed rules to distinguish the GR from the NG points. However, due to the lack of general criteria, such methods usually have difficulties in dealing with point clouds in mixed scenes, and make critical misclassifications when directly processing the whole area. (2) \textit{Learning-based methods}. Different from rule-based methods, the approaches based on hand-crafted features and classifiers first convert ground filtering to a binary classification problem, and then learn the discrimination rules directly from the features. However, the final classification performance is often limited by the low-level hand-crafted features. Compared with hand-crafted features, the high-level features learned with deep neural networks usually have better performance in classification tasks. However, due to the lack of large-scale public datasets, the development of deep-learning-based ground filtering methods is limited.

\subsection{Representative Baseline Methods}

We employ five representative methods with open source software or codes, including two popular rule-based GF methods and three state-of-the-art 3D semantic segmentation networks (SSNs), as solid baselines to benchmark our OpenGF dataset. 

\textit{\textbf{PMF}}~\cite{zhang2003progressive} filters the NG points from the original point cloud through morphological operations (e.g., erosion and dilation) within a series of local windows that progressively change in size. 

\textit{\textbf{CSF}}~\cite{zhang2016easy} first approximates the ground surface with simulated cloth, then extracts GR points from the original point cloud by comparing the unfiltered points and the generated surface.

\textit{\textbf{PointNet++}}~\cite{qi2017pointnet++} is one of the first deep neural network architectures on semantic segmentation built upon PointNet~\cite{qi2017pointnet}, the pioneering work in deep learning-based point cloud processing.

\textit{\textbf{KPConv}}~\cite{thomas2019kpconv} is a point convolutional operation that directly operates on point clouds, and the convolution weights are determined by closeness to defined kernel points in the Euclidean space. The semantic segmentation network KPFCNN has achieved high rankings in several outdoor point cloud benchmarks.

\textit{\textbf{RandLA-Net}}~\cite{hu2020randla} is an efficient semantic segmentation framework specifically designed for large-scale point clouds. 
It features a light-weight local spatial encoding design, and implements attentive pooling to take advantage of the efficiency of random sampling.

\subsection{Evaluation Metrics}
The traditional evaluation metrics of GF are Type \uppercase\expandafter{\romannumeral1}, Type \uppercase\expandafter{\romannumeral2}, and Total errors~\cite{sithole2004experimental}. In order to facilitate the research in related fields, these traditional metrics are replaced with the equivalent or similar metrics widely used in current classification reserach. Specifically, intersection over union ($IoU$) of each class and overall accuracy ($OA$) are used. The computations are provided as follows:

\begin{equation}
OA=\frac{TP_1+TP_2}{TP_1+FP_1+ TP_2+ FP_2}
\end{equation}

\begin{equation}
IoU_1 = \frac {TP_1}{TP_1 + FP_1 + FP_2}
\end{equation}

\begin{equation}
IoU_2 = \frac {TP_2}{TP_2+FP_2+ FP_1} 
\end{equation}

where $IoU_1$ and $IoU_2$ denote the intersection over union of class $NG$ (label 1) and $GR$ (label 2), respectively. $TP_2$, $FP_2$, $FP_1$, and $TP_1$ are, respectively, the number of GR points that are correctly identified, the number of GR points incorrectly identified as NG points, the number of NG points incorrectly identified as GR points, and the number of NG points correctly identified.

In addition, the Root Mean Square Error (RMSE) of the extracted DEM and reference DEM is adopted as a key evaluation metric.

\begin{equation}
RMSE = \sqrt{\frac{\sum_{i=1}^{N} (E_i-R_i)^{2}}{N}}
\end{equation}

where $E_i$ and $R_i$ respect the effective elevation value of the extracted DEM and reference DEM, respectively.
$N$ is the number of pixels with effective elevation value.
\subsection{Parameters and Configurations}

\textit{\textbf{Rule-based baselines}}. The implementation of PMF in PDAL\footnote{\url{https://pdal.io}} and that of CSF in CloudCompare\footnote{\url{https://www.cloudcompare.org}} are adopted. We tried our best to choose the suitable parameters through multiple fine-tunes.

\textit{\textbf{Learning-based baselines}}. In the data preparation stage, we merged class 0 into class 1, so that outliers are treated as NG points to participate in network training. In addition, the minimum coordinates of the points in each file were offset at (0,0,0) to avoid numerical overflow in the networks. The grid size for downsampling was set as 1m after several attempts. The parameters of the networks\footnote{\url{https://github.com/intel-isl/Open3D-PointNet2-Semantic3D}}\footnote{\url{https://github.com/HuguesTHOMAS/KPConv}}\footnote{\url{https://github.com/QingyongHu/RandLA-Net}} were adjusted respectively according to the downsampling grid size regarding the network structure for Semantic3D. 
During training, the batch sizes of different baseline networks were first set to 4, and increased respectively based on the GPU memory usage. Besides, point shuffling and random rotation along z axis were performed as data augmentation.

The above two kinds of baselines were tested on a workstation with an Intel Core i9-9900K CPU @3.60GHz, 32GB RAM, and an NVIDIA RTX 2080Ti GPU, under Ubuntu 18.04.

Moreover, we generated DEMs with a resolution of 0.5m for the ground truth and predicted GR points using the \textit{"LAS Dataset To Raster"} tool (interpolation type: triangulation and natural neighbor) in ArcGIS Pro\footnote{\url{https://www.esri.com/en-us/arcgis/products/arcgis-pro/overview}}. Besides, the \textit{"Fill nodata"} tool in QGIS\footnote{\url{https://www.qgis.org/en/site/}} was adopted to fill the null values at the boundary of all DEMs.

\begin{figure*}[p]
    \centering
    \includegraphics[width=\linewidth]{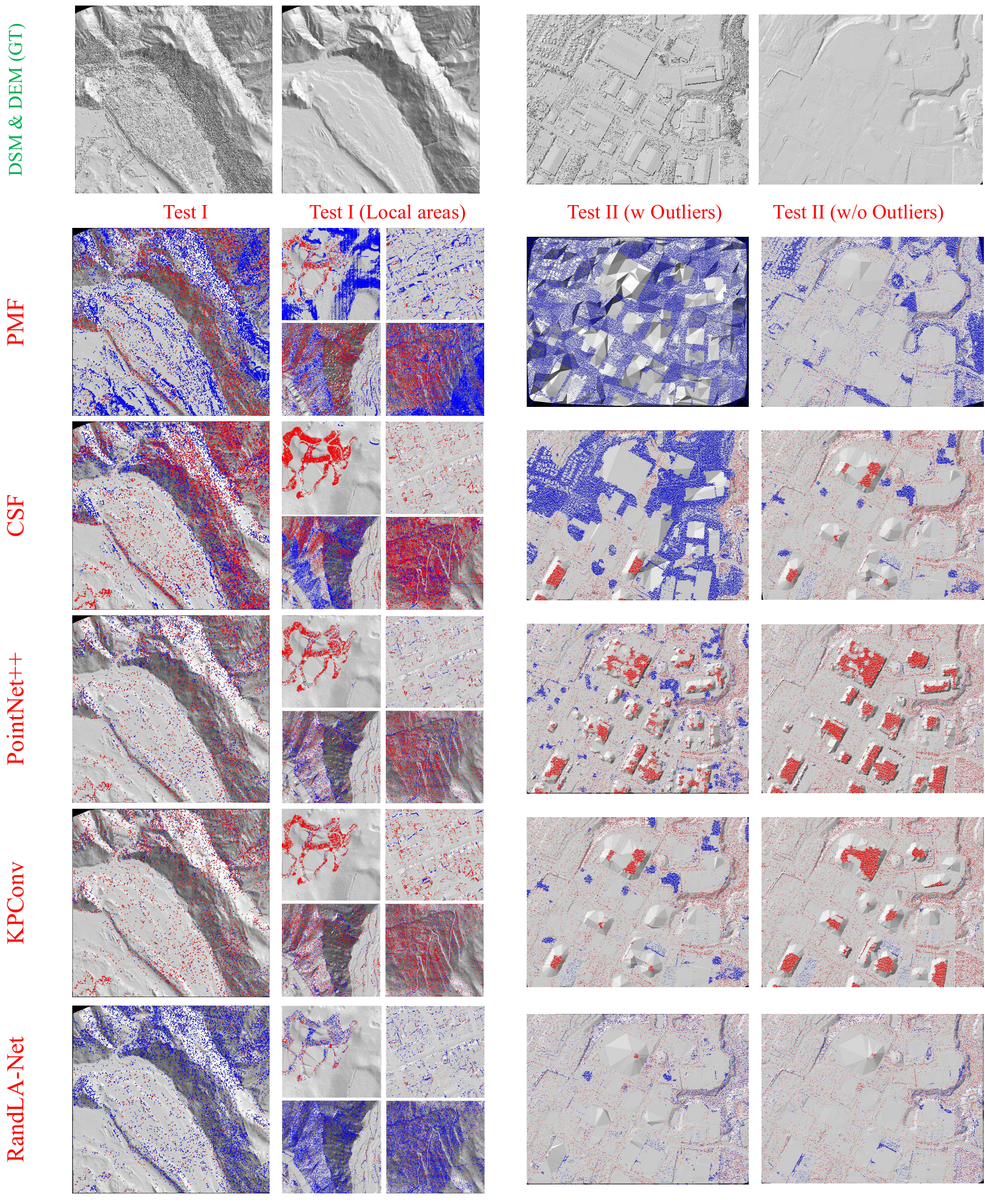}
    \caption{Qualitative results of five baselines on the test set of OpenGF. The top row shows the DSM and DEM (ground truth) of the test data. Red and blue points superimposed on the extracted DEM by different methods represent misclassified NG and GR points, respectively. }
    \label{fig:result}
\end{figure*}  

\begin{table*}[t]
\resizebox{\textwidth}{!}{%
\begin{tabular}{cccccccccccccccc}
\toprule \midrule
\multirow{2}{*}{\textbf{Method}} & \multicolumn{5}{c}{\textbf{Test \uppercase\expandafter{\romannumeral1}}} & \multicolumn{5}{c}{\textbf{Test \uppercase\expandafter{\romannumeral2} (w Outliers)}} & \multicolumn{5}{c}{\textbf{Test \uppercase\expandafter{\romannumeral2} (w/o Outliers)}} \\ \cline{2-16} 
 & $OA$ & $RMSE$ & $IoU_1$ & $IoU_2$ & $ART$ & $OA$ & $RMSE$ & $IoU_1$ & $IoU_2$ & $ART$ & $OA$ & $RMSE$ & $IoU_1$ & $IoU_2$ & $ART$ \\ \hline
PMF & 90.63 & 0.51 & 85.22 & 79.62 & \textbf{1.00} & 50.87 & 66.76 & 50.59 & 1.15 & 1.00 & 86.56 & \textbf{1.11} & 78.50 & 73.61 & 2.00 \\
CSF & 93.07 & 0.95 & 88.17 & 85.64 & 15.00 & 68.36 & 12.24 & 59.94 & 39.90 & 1.00 & 89.34 & 1.89 & 81.08 & 80.38 & 1.00 \\ \hline
PointNet++ & 97.58 & 0.25 & 95.75 & 94.68 & -  & 86.40 & 4.79 & 75.46 & 76.63 & - & 87.38 & 4.89 & 75.19 & 79.63 & - \\
KPConv & \textbf{97.79} & \textbf{0.20} & \textbf{96.10} & \textbf{95.17} & 2.50 & 91.65 & 2.71 & 84.46 & 84.71 & 0.50 & 91.09 & 3.87 & 82.44 & 84.67 & 0.50 \\
RandLA-Net & 96.29 & 0.29 & 93.74 & 91.65 & 2.00 & \textbf{94.28} & \textbf{1.84} & \textbf{89.29} & \textbf{89.05} & \textbf{0.50} & \textbf{94.96} & 1.20 & \textbf{90.38} & \textbf{90.42} & \textbf{0.50} \\\bottomrule
\end{tabular}%
}
\vspace{0.05cm}
\caption{Quantitative results of five selected baselines on our OpenGF dataset. Overall Accuracy (OA, \%), root mean square error (RMSE, meter), per-class IoU (\%) and approximate running time (ART, minute) are reported. ART of PointNet++ is not reported due to low efficiency of selected implementation.}
\label{tbl:result}
\end{table*}

\subsection{Benchmark Results}
Unlike Test \uppercase\expandafter{\romannumeral1}, there are a large number of outliers in Test \uppercase\expandafter{\romannumeral2} (see Figure~\ref{fig:challenges_b}). For comprehensive testing, we transformed the original Test \uppercase\expandafter{\romannumeral2} to Test \uppercase\expandafter{\romannumeral2} (w outliers) and Test \uppercase\expandafter{\romannumeral2} (w/o outliers) by combining outliers into NG and deleting 
outliers respectively. The quantitative results of the five baselines on our test set are listed in Table~\ref{tbl:result}. 
It can be seen that: (1) On Test \uppercase\expandafter{\romannumeral1}, KPConv ranks first in terms of the classification accuracy including the $OA$, $IoU_1$, and $IoU_2$. Its $RMSE$ also surpasses that of all other baselines. PMF costs the minimum time of about 1 minute. (2) On Test \uppercase\expandafter{\romannumeral2} (w outliers), PMF has the worst performance, while RandLA-Net outperforms all other baselines by a substantial margin. Both KPConv and RandLA-Net only 
takes about 0.5 minutes. (3) On Test \uppercase\expandafter{\romannumeral2} (w/o outliers), PMF ranks first in terms of $RMSE$  although its classification accuracy is the worst. Meanwhile, it takes longest time of about 2 minutes. RandLA-Net achieves the best classification accuracy. 

Figure~\ref{fig:result} shows the corresponding qualitative comparison of the results achieved by different baselines. From Figure~\ref{fig:result}, we can intuitively see the amount and distribution of the misclassified NG (red color) and GR (blue color) points. Note that, when we use a smaller grid size (\eg, 0.5 m or 0.2 m) for downsampling in the preprocessing step, just like the result of KPConv, lots of large roof points in Test \uppercase\expandafter{\romannumeral2} are misclassifid as GR points by RandLA-Net.

\section{Comprehensive Analysis}

From above benchmark, we can find the following:
\begin{itemize}[leftmargin=*]
\setlength{\itemsep}{0pt}
\setlength{\parsep}{0pt}
\setlength{\parskip}{0pt}
  \item  State-of-the-art 3D SSNs outperform rule-based GF methods by a substantial margin in terms of the classification accuracy (especially on point clouds in mixed scenes), thereby verifying our assumption that deep learning models with good generalization ability can be trained with the OpenGF dataset. The implicit rules learned from the data are more robust than the manually designed explicit rules in diversified terrain scenes.
  \item  Although state-of-the-art 3D SSNs have achieved impressive classification accuracy, there is still room for reducing the $RMSE$ of DEMs. This mainly attributes to the micro-terrain errors and misclassified large roof points in the extracted GR points. 
  \item  State-of-the-art 3D SSNs are much less sensitive to low outliers than rule-based GF methods. This result is not surprising considering that rule-based GF methods usually assume local lowest points as ground, while 3D SSNs classify each point indifferently.
  \item  Large-scale spatial context is essential for the recognition of large targets. As is expected, 3D SSNs that can only consume inputs with a small coverage at one time failed to accurately identify large roofs. 
   \item  The running time of 3D SSNs increases linearly on the amount of data, while that of rule-based GF methods is highly dependent on the parameter settings and the amount of data. 
\end{itemize}

\section{Conclusion and Outlook}

This paper presents OpenGF, the first ultra-large-scale point cloud dataset with high-quality GR annotations dedicated for GF. The dataset not only covers approximately 47.7 $km^2$ with over 542 million points from 4 countries, but also contains 9 different terrain scenes. Two popular rule-based GF algorithms and three state-of-the-art 3D SSNs are selected as baselines. Through extensive benchmarking and comprehensive analysis, we argue that deep learning models for GF can be effectively trained on OpenGF, and the effective ways should be further explored for the correction of micro-terrain errors and the removal of large roofs. 

In the near future, our OpenGF dataset and benchmark may be a stepping stone towards advancing research in related areas. \eg, (1) the deep learning models pre-trained on OpenGF can be quickly fine-tuned to filter photogrammetric point clouds via transfer learning. (2) the OpenGF dataset can be further labeled and expanded into an ultra-large-scale semantic segmentation dataset containing diverse terrain scenes.

\section*{Acknowledgments}
This work was supported in part by the National Natural Science Foundation of China (42001400).
OpenGF contains information licensed under the Open Government Licence – Ontario, the CC BY 4.0 \& the CC BY-SA 4.0 Licence. Part of this work is based on [data, processing] services provided by the OpenTopography Facility with support from the National Science Foundation under NSF Award Numbers 1948997, 1948994 \& 1948857.

{\small
\bibliographystyle{ieee_fullname}
\bibliography{references}
}

\end{document}